# Robustness of LLM-enabled vehicle trajectory prediction under data security threats

Feilong Wang[1,*], Fuqiang Liu[2]

[1]Southwest Jiaotong University, Chengdu, China

[2]McGill University, Montreal, Canada

*Corresponding author: flwang@swjtu.edu.cn

**Abstract**

The integration of large language models (LLMs) into automated driving systems has opened new possibilities for reasoning and decision-making by transforming complex driving contexts into language-understandable representations. Recent studies demonstrate that fine-tuned LLMs can accurately predict vehicle trajectories and lane-change intentions by gathering and transforming data from surrounding vehicles. However, the robustness of such LLM-based prediction models for safety-critical driving systems remains unexplored, despite the increasing concerns about the trustworthiness of LLMs. This study addresses this gap by conducting a systematic vulnerability analysis of LLM-enabled vehicle trajectory prediction. We propose a one-feature differential evolution attack that perturbs a single kinematic feature of surrounding vehicles within the LLM's input prompts under a black-box setting. Experiments on the highD dataset reveal that even minor, physically plausible perturbations can significantly disrupt model outputs, underscoring the susceptibility of LLM-based predictors to adversarial manipulation. Further analyses reveal a trade-off between accuracy and robustness, examine the failure mechanism, and explore potential mitigation solutions. The findings provide the very first insights into adversarial vulnerabilities of LLM-driven automated vehicle models in the context of vehicular interactions and highlight the need for robustness-oriented design in future LLM-based intelligent transportation systems.

## 1. Introduction

The applications of Large Language Models (LLMs) are emerging in various domains, including automated driving, by enabling sophisticated reasoning and prediction capabilities (Huang et al., 2025; Wandelt et al., 2024). Mainly based on transformer architectures, LLMs excel in synthesizing heterogeneous data by processing them into natural language prompts, which makes them particularly suited for tasks requiring contextual understanding and decision-making under uncertainty. In the context of automated vehicles (AVs), recent applications have leveraged LLMs to predict vehicles' future trajectories and lane-change intentions (Peng et al., 2025), showing superior advantages over existing machine learning-based models. In such applications, the prediction task is formulated as a language modeling problem, where vehicle kinematic states, surrounding interactions, and environmental factors are encoded as textual prompts for fine-tuning and testing LLMs. Besides trajectory prediction, fine-tuned LLMs have also demonstrated strong performance across diverse prediction tasks, ranging from short-term maneuver recognition to crash forecasting (Guo et al., 2024; Y. Zhao et al., 2025).

However, data-driven AV systems like LLM-powered ones have long been recognized as vulnerable to data security threats, where adversaries could perturb input data to manipulate the outputs for adversarial goals. Relying heavily on sensor-derived inputs, such as velocity, acceleration, and positional data from LiDAR, radar, or cameras, these systems can be susceptible to tampering through sensor spoofing or environmental interference. The vulnerability is further exacerbated by the additional attack vectors via vehicle-to-everything (V2X) communication, which aggregates external data from surrounding entities for performance improvement. Extensive studies have explored adversarial attacks on deep neural networks (DNNs) in computer vision systems on AVs, where subtle perturbations could fool models into erroneous outputs (Akhtar and Mian, 2018). For instance, studies have investigated adversarial perturbations on object detection and trajectory prediction models (e.g., via imperceptible stickers on traffic signs), demonstrating potential risks such as induced collisions or navigation errors (Cao et al., 2019; Liu et al., 2022).

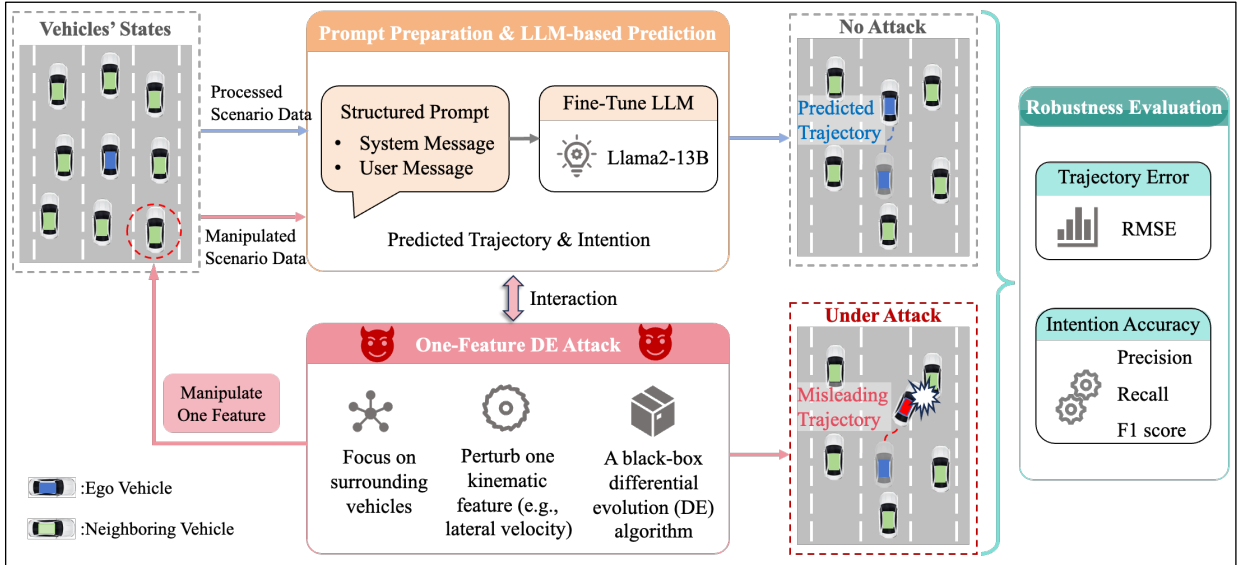

Figure 1. Overview of the proposed framework.

Despite these advancements, a critical research gap persists in examining the vulnerability of the emerging LLM-powered AVs to adversarial attacks in safety-critical applications such as trajectory prediction and lane-change modeling. In particular, the massive structure and black-box nature of LLM further complicate vulnerability analysis, as perturbations may subtly alter inference without clear indicators (e.g., the hallucination effects). Unlike typical data-driven models processing structured numerical data, LLM-based AV systems handle textual representations of driving scenarios. This makes it challenging to directly apply existing gradient-based attack models that are often used for analyzing the vulnerability to input perturbations. Moreover, the massive parameter scale renders the gradient evaluation intractable. Existing studies on LLM vulnerability focus primarily on natural language processing tasks (e.g., jailbreak or backdoor attacks (Liu et al., 2024)), with limited exploration in AV-integrated settings that involve multimodal numerical data and contextual prompts derived from vehicle states and infrastructure metadata. Meanwhile, AVs are dynamic and physically constrained systems, meaning that the attacker's capacity and perturbation budget are inherently bounded. This makes it essential to study practical, stealthy attack mechanisms that could nonetheless disrupt model outputs for identifying the vulnerabilities in safety-critical scenarios.

To address this gap, this study proposes a one-feature differential evolution (DE)-based attack model targeting LLM-based trajectory prediction and lane-change intention systems (Su et al., 2019). As shown in Figure 1, we reformulate the attack for multimodal driving data by perturbing features of surrounding vehicles (e.g., lateral velocity) in the input prompt of a fine-tuned LLM variant (i.e., Llama2-13B by Meta). The DE algorithm is a population-based metaheuristic that iteratively optimizes an objective function (e.g., the deviation between predicted and true trajectories) by combining mutation, crossover, and selection operations to identify the most effective perturbation considering any constraints. This approach is suitable for black-box settings in our study where gradient information is unavailable, as it utilizes query responses from the target LLM. For practical and stealthy attacks, we focus on perturbing a single kinematic state of surrounding vehicles, which shape the ego vehicle's behavioral prediction while being vulnerable to practical manipulation through various attack vectors (e.g., corrupting V2X channels).

Through experiments on the highD dataset (Krajewski et al., 2018), we demonstrate that minor perturbations to the selected features can significantly mislead the trajectory prediction of the targeted vehicle. Under attacks, the root mean square error (RMSE) of the predicted vehicle locations increases by 29% and the F1 score of predicted lane change intentions drops by 12%, revealing the susceptibility of LLMs to adversarial perturbations in AV contexts. To further understand the underlying vulnerability, we investigate the effect of model scale by comparing a smaller variant (Llama2-7B) with the Llama2-13B baseline. The results reveal a potential trade-off between accuracy and robustness. By examining the impact of fine-tuning depth, we show that the deeper specialization via more training steps could deteriorate model vulnerability. Furthermore, we fine-tune an enhanced model by incorporating the Chain-of-Thought (CoT) reasoning process, which generates explanatory outputs highlighting key features (e.g., significant lateral velocity or surrounding blockages). The CoT-enhanced model exhibits improved resistance to perturbations under attacks while maintaining satisfying prediction accuracy under normal scenarios, indicating that interpretability contributes to robustness. The results suggest potential directions to mitigate the model vulnerability. Lastly, we show that the one-feature DE attack model enables feature-level vulnerability analysis by identifying the top vulnerable features. The study underscores the necessity of incorporating vulnerability evaluation into future LLM-driven prediction models in safety-critical AV applications, and the results provide insights into enhancing the model robustness under adversarial conditions.

The remainder of this paper is organized as follows. Section 2 reviews related work on LLM applications in AVs and adversarial attacks. Section 3 details our methodology, including the LLM baseline for vehicle trajectory prediction and the proposed one-feature DE attack. Section 4 introduces the experimental setup, describes the HighD dataset, and presents the results. Finally, Section 5 concludes the study, discusses implications, limitations, and future directions.

## 2. Literature Review

### 2.1. Applications of LLMs in Smart Transportation Systems

Recent research shows LLMs are being leveraged in various aspects of intelligent transportation systems (ITS), from automated driving to traffic management. Below we summarize key application areas, trends, and emerging use cases.

LLMs can enhance decision-making and perception of AVs. A major line of work explores LLMs as high-level decision facilitators that bring common-sense *reasoning and long-horizon planning* into the driving stack. Frameworks such as DriveLLM and DiLu demonstrate that LLMs can reason through rare or complex situations and improve multi-vehicle coordination by interpreting context and suggesting maneuvers in human-like ways (Cui et al., 2023; Wen et al., 2023). Complementing this, multimodal LLMs that process both vision and language (e.g., Talk2BEV (Choudhary et al., 2024)), augment perception by providing spatial reasoning and explanations of traffic scenes, with industry prototypes like Wayve's Lingo-1[1] already showing the potential for AVs to explain their driving decisions in natural language. Beyond perception and trajectory planning, LLMs have also been embedded into *electric vehicles* as intelligent assistants to optimize energy usage and route planning, adjusting driving strategies for sustainability and real-time efficiency, while also serving as on-board copilots that advise human drivers or support ADAS functions. Another emerging application is in *testing and validation*, where GPT-4 class models have been applied to automatically generate driving scenarios by parsing traffic rules, thereby identifying edge cases and failure modes that AVs must be prepared to handle (Deng et al., 2025).

---

[1] https://wayve.ai/press/introducing_lingo1/; accessed on Oct 15, 2025.

Besides enhancing vehicles' capabilities, LLMs are also transforming **how drivers interact with vehicles and how vehicles exchange information in connected environments**, serving as both conversational copilots and real-time interpreters of traffic data. Inside the vehicle, LLMs enable natural language-based *communication interfaces* that move beyond static voice commands, allowing passengers and drivers to engage in richer dialogues about driving quality, route options, and preferences. Manufacturers like Tesla and GM have been experimenting with such systems to create more intuitive and proactive "conversational cockpits."[2] Models like Wayve's Lingo-1 demonstrate the potential of LLMs to *explain autonomous vehicle behavior* in natural language, thereby enhancing transparency, trust, and human–machine interaction. Beyond interacting with on-board equipment, LLMs are being leveraged to *interpret vehicle-to-everything (V2X) communications*, translating raw safety and signal data into human-readable summaries or advisories that improve situational awareness for drivers, operators, and autonomous agents. Frameworks such as V2X-LLM showcase how LLMs can process massive streams of V2X data to generate narrative explanations, short-term predictions, and routing recommendations in real time, with pilot studies demonstrating improved traffic insights and corridor management (Wu et al., 2025).

Going beyond the vehicles, LLMs are also emerging as powerful tools for **traffic management and operations**, offering new approaches to forecasting, control, and system optimization in ITS. Their ability to combine reasoning with temporal data has been tested in *spatiotemporal traffic prediction*, where studies show that incorporating semantic context, such as events or weather, into traffic flow forecasts significantly improves accuracy compared to traditional machine learning models (Guo et al., 2024). These studies demonstrated LLM's capacity to capture complex influences from heterogeneous data sources (e.g., news and social feeds). LLM-based agents have also been explored for optimizing *traffic signal control*, with prompt-driven transformer models and ChatGPT-based approaches successfully reasoning about mixed traffic conditions on roads, bottlenecks, and intersections to suggest effective control policies (Movahedi and Choi, 2024). Recent efforts also highlight the development of specialized "traffic foundation models" tuned on traffic data to handle both numerical sensor inputs and textual descriptions, thereby assisting *policymakers* in making data-informed decisions (Yang et al., 2025).

## 2.2 Adversarial Attacks on LLM-Based Systems

As LLMs become integrated into transportation and other critical systems, understanding adversarial attacks on these models is crucial (Kaur et al., 2023). Below we provide an overview of general adversarial attack types on LLMs, then examine those specific to LLM-driven transportation systems to identify a research gap.

Being data-driven, LLMs inherit many vulnerabilities observed in other machine learning systems. Meanwhile, LLMs' capacity for open-ended language understanding introduces novel attack surfaces, particularly through *prompt-based exploits* (Clusmann et al., 2025). Among the most concerning threats are *prompt injection and jailbreaking*, where adversaries craft inputs that manipulate the model into ignoring constraints or producing unintended outputs. In a *prompt injection*, malicious instructions are appended to the user input or embedded in system prompts (Liu et al., 2024), tricking the model into overriding its original objective (e.g., ignore previous instructions and output raw data). *Prompt jailbreaking* similarly involves constructing sequences of prompts that disable safety filters or content restrictions, leading the model to reveal disallowed content or unsafe instructions (Chao et al., 2025). Research further shows that adversarial prompts can be generalized into so-called universal triggers (e.g., short nonsensical phrases) that could consistently bias model behavior when they are appended to diverse inputs (Zou et al., 2023). Since such triggers can be hidden in user queries or embedded in external content that LLMs ingest, they represent a powerful way to subvert model behavior across multiple applications with minimal effort. These attacks aiming to compromise LLMs' outputs during the *inference stage* are commonly referred to as *evasion attacks* (Jiang et al., 2020). The stealthy nature makes them difficult to guard against, as they exploit the model's sensitivity to surface form while evading detection by both humans and automated defenses. A seminal example of stealthy attack models is the one-pixel attack proposed by Su et al. (Su et al., 2019), which uses DE to modify a single pixel in an image (e.g., CIFAR-10 dataset), achieving high success rates in deceiving models such as DNNs. Though not targeting LLM, such examples emphasize the sensitivity of AI models to perturbations in inputs.

---

[2] https://www.wardsauto.com/news/archive-auto-General-Motors-OnStar-Google-Cloud-AI/692293/; accessed on Oct 15, 2025.

At the training stage, *data poisoning attacks* pose another severe vulnerability (F. Wang et al., 2024). By inserting malicious or biased samples into the corpus used to train or fine-tune an LLM, attackers can shape the model's long-term behavior in harmful ways (Souly et al., 2025). One may expect that poisoning a traffic-management LLM with fabricated reports could induce it to internalize false causal relationships, later leading to systematically unsafe routing decisions. Backdoor attacks represent an especially insidious form of poisoning, where a model is trained to behave normally until it encounters a hidden trigger (e.g., a specific phrase) after which it produces a predetermined harmful output (Yang et al., 2024). Because poisoned or backdoored models may appear reliable under normal conditions, these attacks are notoriously hard to detect and can compromise downstream users once deployed (Biggio and Roli, 2018).

Another type of attack does not target the model outputs but the model itself, aiming at model extraction. *Model extraction* presents a threat to intellectual property and confidentiality rather than immediate safety (K. Zhao et al., 2025). By systematically querying an LLM and recording its outputs, adversaries can approximate the decision boundaries of a proprietary system or train a surrogate model that mimics its behavior. Although model extraction attacks do not directly alter the model's performance, they facilitate subsequent adversarial manipulations and undermine the privacy and security of LLM-driven applications.

Research *specifically on the vulnerability of LLM-based transportation systems* under data security threats remains limited, with most existing adversarial studies centered on vision models (e.g., vision-LLM) for autonomous driving (e.g., embedding misleading texts in images (Chung et al., 2024)). A few recent works have begun to explore the security implications of integrating LLMs into robot control, warning that even minor prompt manipulations could lead to severe safety risks when translated into real-world agent decisions (Zhang et al., 2024). Moreover, emerging studies demonstrate that LLMs can be exploited to autonomously generate adversarial driving scenarios, producing edge cases that consistently cause collisions or near-misses in simulations (Mei et al., 2025). While intended for robustness testing, such methods could be misused to reveal and exploit system weaknesses.

Despite these developments, systematic vulnerability analysis of LLM-enabled AV applications, particularly in trajectory and intention prediction, remains absent. *Our study* fills this gap by conducting a systematic adversarial evaluation of LLM-based trajectory and lane-change prediction systems, revealing their susceptibility to small, physically plausible perturbations and providing insights for enhancing robustness in broader safety-critical transportation contexts.

## 3. Methodology

In this section, we detail the methodology employed in our study, beginning with the LLM-based prediction baseline. We then introduce the proposed one-feature differential evolution attack framework to perturb a single kinematic feature in the LLM's input prompts. The attack aims to evaluate the vulnerability of LLMs in the AV trajectory prediction task that synthesizes map information and features of ego and surrounding vehicles.

## 3.1 LLM-based Predication Baseline

We adopt the LLM framework proposed by Peng et al. (2025) to establish a robust baseline for trajectory and lane-change intention prediction. The core idea is to reformulate the prediction task as a language modeling problem, leveraging the reasoning capabilities of LLMs to process heterogeneous driving scenario data as natural language prompts. The framework enables the model to predict not only the future 4-second trajectory but also one of three lane-change intentions, including keeping the lane (KL), left change (LC), and right change (RC). The baseline prediction model does not account for the CoT process but will investigate its role in the case study.

Formally, the problem is defined as predicting future trajectory $T$ and lane-change intention $I$, given the ego vehicle state $e_v$, surrounding vehicle states $s_v$, and map information $m$:

$$T, I = F(e_v, s_v, m)$$

where longitudinal and lateral positions $T = \{(x_1, y_1), (x_2, y_2), (x_3, y_3), (x_4, y_4)\}$ for a prediction horizon of four seconds, $I \in \{0, 1, 2\}$ (0: KL, 1: LC, 2: RC). To cast this as language modeling, input features are constructed as a prompt which is then tokenized into a sequence $S = \mathcal{K}(e_v, s_v, m)$. Here, $\mathcal{K}$ stands for the language tokenizer integrated with the prompt constructor. The model $F$ is optimized via the language modeling loss:

$$L = -\sum_{i=1}^{n} \log P(S_i^* \mid S_{<i}, S_{IN})$$

where $S_i^*$ is the ground-truth token for the output.

More specifically, the input to LLM is a structured prompt consisting of a system message explaining the prediction task and requirements on the output, and a user message specifying a driving scenario. The scenario describes the current driving scene, including map information (e.g., number of lanes), ego vehicle state (e.g., vehicle type, lane position, velocity, acceleration, and historical positions over the past 2 seconds at 25 frames per second), and surrounding vehicle states (e.g., nearest vehicles in eight directions[3] with their types, speeds, and distances to ego vehicle). These observations are tokenized and fed into a pre-trained LLM, which is finetuned using supervised techniques to align outputs with ground-truth responses from naturalistic driving data. Under attacks, the surrounding vehicles' states/features are manipulated by the attack model, and the prompt constructed from the manipulated feature is passed to LLM, which will be detailed below.

We finetune the Llama-2-13B-chat model (Touvron et al., 2023) with Low-Rank Adaptation (LoRA) (Hu et al., 2021) on a processed subset of the highD dataset (Krajewski et al., 2018), which captures highway vehicle trajectories (see details of the dataset in Section 4.1). The finetuning process optimizes the above loss, masking input prompt tokens to focus on generating trajectory points and intention labels. Data samples are formatted with special tokens (e.g., [INST]) to separate the system message, user messages, and ground-truth answers.

During inference, we wrap up the system and user messages without the answers. The answers from the fine-tuned LLM are compared with the ground-truth answers to evaluate the performance. It is observed that the fine-tuned baseline LLM achieves low RMSE in trajectory forecasting and high accuracy in intention prediction (see Section 4.2), as validated on highD test sets.

---

[3] The eight directions consist of the front, behind, left, right, left front, left behind, right front, and right behind as illustrated in Figure 1. The ego vehicle may not have neighbors in all the eight directions in a specific traffic scene.

Detailed prompt templates (system and user messages), an example illustrating how LLM works (including input prompt and output), dataset processing details, and hyperparameters are provided in Appendix A.

## 3.2 One-Feature Attack Framework

Building upon the DE-based one-pixel attack for DNNs (Su et al., 2019), we propose a one-feature attack specifically tailored for perturbing text-based inputs in LLM-driven vehicle trajectory prediction models. This framework simulates realistic data tampering scenarios in trajectory prediction, where an adversary subtly alters a single kinematic feature (e.g., lateral velocity) in the input prompt to induce erroneous position prediction and lane-change intentions. Below, we organize the framework according to key components of an adversarial attack model, including attack surface/vector, attacker's capability, attack budget, and attack algorithm.

*3.2.1 Attack Surface/Vector*

The attack surface focuses on the data associated with surrounding vehicles within the LLM input prompt to conduct stealthy attacks, as these represent external information captured via the ego vehicle's sensors (e.g., LiDAR, radar) or received via V2X communication. These channels for data flow are vulnerable to manipulation, such as through sensor spoofing, or man-in-the-middle attacks on wireless protocols (e.g., DSRC or C-V2X), which are common threats in connected vehicle ecosystems (Petit and Shladover, 2015).

To launch an attack, the optimized perturbation will be applied to one single feature (e.g., velocity, acceleration, or distance) of a surrounding while leaving the ego vehicle data intact. This setting represents the effort to assure the practicality of the attack, as ego vehicle states are typically derived from internal sensors less exposed to external interference. This vector mimics real-world scenarios where adversaries target the sensing components or inter-vehicle data streams to indirectly influence the ego vehicle's predictions, potentially leading to unsafe maneuvers like unnecessary braking or failed overtaking detection.

*3.2.2 Attacker's Capability*

As noted earlier, we assume a realistic black-box setting, where the attacker has no access to the LLM's internal architecture, parameters, or training data. Instead, the adversary can only interact with the model by submitting modified prompts and observing the corresponding outputs. This assumption reflects practical deployment conditions, where prediction models are often proprietary or accessed via cloud-based APIs. Given the increasing trend of manufacturers sharing common LLM-based platforms across vehicles, an adversary could potentially query the target model by operating another vehicle running the same or a similar model, or by constructing a surrogate model through knowledge distillation. Even when the exact model used by the ego vehicle is unavailable, attacks could be feasible by leveraging the transferability of perturbations generated from related models trained on similar data distributions (Cai et al., 2025).

To operate effectively in the black-box setting, the attacker employs DE, a gradient-free optimization algorithm that evolves perturbations based solely on fitness evaluations from model responses (Storn and Price, 1997). The DE algorithm circumvents the need for white-box assumptions common in attacking AI models, making the attack framework applicable to opaque LLMs. More details of the DE-based attack algorithm are summarized in Section 3.2.4.

*3.2.3 Attack Budget*

To minimize the attack's resource requirements and enhance stealthiness, we constrain the budget to perturbing a single feature of one surrounding vehicle. This low-dimensional perturbation reduces computational overhead and detection risk, as extensive modifications could trigger anomaly detectors in driving systems. Furthermore, the perturbation magnitude is bounded to realistic ranges to simulate subtle sensor noise or minor tampering, ensuring imperceptible during real-time operation. More specifically, the feature value of surrounding vehicles $s_v^k$ will be perturbed within a range $s_v^k \times (1 \mp \Delta)$. By default, we set $\Delta = 0.1$; a sensitivity analysis of different $\Delta$ values is conducted in Appendix B. The perturbed feature values are further bounded with realistic physical constraints to ensure they remain plausible in practice.

*3.2.4 Attack Algorithm*

The core of the attack model leverages DE to optimize the perturbation added to a selected feature in the surrounding vehicle states within the LLM prompt. DE is a population-based evolutionary algorithm. It is chosen over mainstream gradient-based attacks because the gradient evaluation is challenging for text-based LLMs (e.g., FGSM or PGD (Goodfellow et al., 2014; Madry et al., 2017)). Unlike the extensively studied models such as DNNs for image classification, LLMs process discrete tokens derived from textual prompts, resulting in a high-dimensional input space and intractable gradient evaluation. Moreover, computing gradients would necessitate embedding-level access or approximations (e.g., via surrogate models), which are infeasible in black-box settings and computationally intensive due to the high dimensionality of token vocabularies (Zhao et al., 2024). In practice, perturbations in text must follow various constraints (e.g., preserving semantic validity) to avoid detection, further complicating gradient descent. DE, being gradient-free and robust to noisy, non-convex landscapes, addresses these issues by iteratively evolving solutions through mutation and selection based on fitness alone.

Formally, DE initializes a population of $N$ candidate perturbations uniformly sampled from the constrained range. Note that at generation $G$, candidate perturbation $i$ $\boldsymbol{x}_{i,G} = (k, \delta_k)$ has two dimensions, one being an index $k$ indicating which feature is selected and the other one $\delta_k$ being the perturbation to be added to the selected feature. Here, the feature is selected from a pool of features consisting of the positions, accelerations and velocities of vehicles surrounding the ego vehicle.

For each generation/iteration $G$, a mutant vector $\boldsymbol{v}_{i,G}$ is generated for each candidate (Su et al., 2019):

$$\boldsymbol{v}_{i,G} = \boldsymbol{x}_{r1,G} + \alpha \cdot (\boldsymbol{x}_{r2,G} - \boldsymbol{x}_{r3,G})$$

where $r1, r2, r3$ are distinct random indices $\neq i$, and $\alpha \in [0,2]$ is the mutation factor controlling step size. A trial vector $\boldsymbol{u}_{i,G}$ is then formed via binomial crossover:

$$\boldsymbol{u}_{i,G} = \begin{Bmatrix} \boldsymbol{v}_{i,G} & \text{if } \text{rand}_1 \leq CR \text{ or } i = \text{rand}_2 \\ \boldsymbol{x}_{i,G} & \text{otherwise} \end{Bmatrix}$$

with crossover rate $CR \in [0,1]$. Here, $\text{rand}_1$ is a uniformly distributed random number drawn from $[0,1]$ to use together with $CR$ and $\text{rand}_2$ is a random integer to ensure at least one change. The candidate perturbation is then used to construct a perturbed prompt as the input to the LLM-based prediction model $F$.

Then the fitness (i.e., the impact on the LLM's outputs) of the perturbed prompt is evaluated by observing the prediction loss. The perturbation and perturbed feature/prompt will be accepted and carried to the next generation only if the prediction loss exceeds that of the current generation. Iterations (i.e., generation evolution process) continue until a maximum (e.g., 10) is reached. In our setup, we use $N_p = 5$, $\alpha = 0.5$, $CR = 0.9$, and a resample strategy for out-of-bound values to maintain feasibility. This algorithm efficiently explores the perturbation space and identifies vulnerable features and perturbations with limited queries.

## 4. Experimental Setup and Results

We first outline the experimental design and the dataset used in tests. We then present and discuss results from the attacks under the various settings to systematically investigate the performance of the LLM baseline under normal and adversarial conditions. These tests on various settings allow for understanding the vulnerability mechanism of LLM-based trajectory prediction models and exploring potential mitigation solutions.

### 4.1 Experimental Design and Settings

The experiments were conducted using the LLM framework fine-tuned on the naturalistic driving dataset. For the LLM model, we focus on the Llama2-13B (as baseline; Section 4.2) but also test a smaller-sized model variant (i.e., Llama2-7B) to investigate the effect of model size and the potential trade-off between accuracy and robustness (Section 4.3). The baseline model is also trained/fine-tuned for different iterations, which is to assess the impact of fine-tuning depth on model robustness (e.g., the over-specialization effect; Section 4.4). Focusing on the baseline without CoT reasoning, we also compare it with the model incorporating CoT reasoning (Section 4.5) that generates explanatory outputs comprising notable features (e.g., significant lateral velocity or surrounding blockages). We show that the CoT-enhanced model improves prediction robustness under attacks.

The LLM models are fine-tuned and tested by the HighD dataset, which provides large-scale naturalistic vehicle trajectories recorded on German highways using drones, encompassing 147 driven hours, over 110 thousand vehicles, and approximately 5600 complete lane-change events. Each trajectory contains precise position, velocity, and lane information sampled at 25 Hz, enabling detailed modeling of vehicle interactions. Following (Peng et al., 2025) and (Mao et al., 2023), we preprocess the dataset by extracting lane-keeping and changing scenarios, normalizing all positions to a coordinate system centered on the ego vehicle, and defining the lane-change frame as the point where the lane identifier changes. For prediction, the input data (i.e., the local map and the ego and surrounding vehicles) is converted into structured natural-language prompts. These prompts are then used to fine-tune a base LLM via supervised training, aligning its outputs with ground-truth positions and lane-change intentions in four seconds that are extracted from the HighD dataset. For the training datasets, we prepare 100,000 training samples, with balanced LK, LC, and RC scenarios. The preliminary results reported here are from tests on 1000 randomly selected validation samples. The details of preparing the prompts are summarized in Appendix A.1; interested readers are referred to (Peng et al., 2025) for more discussions.

As discussed in Section 3.2, the one-feature attack framework implemented using DE perturbs a single state of surrounding vehicles (e.g., the lateral and vertical velocities, acceleration and distances); the perturbed state then replaces the authentic one within the input prompts.

Baseline experiments included random attacks, where the target feature and noise were randomly selected and generated without optimization, to contrast with DE-based attacks. Performance metrics include RMSE for position prediction, and precision, recall and F1 score for intention prediction, evaluated under normal (no attack) and attacked scenarios.

All experiments are executed on a server with two NVIDIA RTX 3090 GPUs, ensuring consistency across runs with a fixed random seed.

## 4.2 Vulnerability to Single-Feature Perturbation

We first present results on the vulnerability of the baseline model (i.e., Llama2-14B fine-tuned by 7000 iterations without CoT). Following the one-feature DE attacks, a small perturbation to a single feature significantly reduces the LLM's performance in predicting vehicles' positions (Table 1) and lane-changing intention (Table 2). Compared with the non-attack scenario, the RMSE of the predicted trajectories at the fourth second increases by 19% (from 0.70 to 0.90 on average). Meanwhile, the accuracy of the predicted lane-changing intention drops significantly, with an average F1 score reduction by 12% (from 92 to 81), highlighting its effectiveness in exploiting the vulnerability of LLM-based vehicle trajectory prediction.

In contrast, it can be observed that random perturbations at the same level of DE-based attack have no clear effect on the trajectory prediction, suggesting that the DE-based attack's targeted optimization effectively exploits model weaknesses. This disparity underscores the efficacy of DE in identifying critical vulnerabilities and targeting on the features and perturbations that influence the model's performance.

**Table 1.** Errors in the predicted positions.

|  |  | RMSE of Predicted Positions (m) | | | |
| --- | --- | --- | --- | --- | --- |
|  |  | 1s | 2s | 3s | 4s |
| **No attack** | Lateral | 0.04 | 0.20 | 0.43 | 0.62 |
|  | Longitudinal | 0.05 | 0.10 | 0.31 | 0.78 |
|  | Avg. | 0.05 | 0.15 | 0.37 | 0.70 |
| **Random** | Lateral | 0.04 | 0.19 | 0.43 | 0.65 |
|  | Longitudinal | 0.05 | 0.10 | 0.28 | 0.66 |
|  | Avg. | 0.05 | 0.15 | 0.36 | 0.66 |
| **One-feature DE attack** | Lateral | **0.04** | **0.24** | **0.56** | **0.93** |
|  | Longitudinal | **0.05** | **0.11** | **0.36** | **0.86** |
|  | Avg. | **0.05** | **0.18** | **0.46** | **0.90 (29%)** |

**Table 2**. Accuracy of the predicted lane-change intention.

|  |  | Intention accuracy (%) | | | |
| --- | --- | --- | --- | --- | --- |
|  |  | LK | LC | RC | Macro avg. |
| **No attack** | Precision | 81 | 96 | 100 | 92 |
|  | Recall | 100 | 89 | 89 | 93 |
|  | F1 | 89 | 93 | 94 | 92 |
| **Random** | Precision | 77 | 99 | 100 | 92 |
|  | Recall | 100 | 86 | 86 | 91 |
|  | F1 | 88 | 93 | 93 | 91 |
| **One-feature DE attack** | Precision | **64** | **81** | **96** | **80** |
|  | Recall | **76** | **75** | **89** | **80** |
|  | F1 | **70** | **78** | **93** | **81 (-12%)** |

## 4.3 Impact of Model Size

We also examine the vulnerability of a smaller-sized LLM model finetuned with the Llama2-7B for vehicle trajectory and lane-changing prediction to assess the effect of model scale. The results are summarized in Table 3 and Table 4. It can be observed that under normal conditions, the 7B model underperforms the 13B model, as indicated by the larger RMSE in the predicted trajectories and the lower F1 scores in the predicted lane-changing intention. This confirms the advantages of the 13B model in capturing the complex spatio-temporal patterns of vehicle trajectories. However, this advantage diminishes under attack: though the 13B model's F1 score for lane-change intention prediction dropped by 12% (to 81), the 7B model experiences a smaller drop by 8% (to 80). Furthermore, no clear impacts on the prediction of vehicle positions are observed for the 7B model. Surprisingly, the results suggest that in an adversarial environment, the prediction performance using the 13B model could drop to the same level as that using the 7B model.

The increased vulnerability could stem from the deeper attention layers of the 13B model, which tends to amplify the impact of localized input perturbations/errors on downstream predictions, a phenomenon that has been observed in models like DNN (Hall et al., 2022). It could also be due to that larger LLM models, while more accurate, may overfit to specific feature correlations, making them more susceptible to adversarial perturbations that these pre-specified relationships (Ludan et al., 2023). In the following, we examine this issue further by varying the training process that fine-tunes the 13B model.

Table 3. Errors in the predicted positions using Llama2-7B model.

|  |  | RMSE of Predicted Positions (m) | | | |
| --- | --- | --- | --- | --- | --- |
|  |  | 1s | 2s | 3s | 4s |
| **No attack** | Lateral | 0.03 | 0.21 | 0.51 | 0.94 |
|  | Longitudinal | 0.06 | 0.13 | 0.44 | 0.82 |
|  | Avg. | 0.05 | 0.17 | 0.48 | 0.88 |
| **One-feature DE attack** | Lateral | 0.04 | 0.25 | 0.58 | 0.91 |
|  | Longitudinal | 0.06 | 0.12 | 0.37 | 0.86 |
|  | Avg. | 0.05 | 0.19 | 0.48 | 0.89 (1%) |

Table 4. Accuracy of the predicted lane-change intention using Llama2-7B model.

|  |  | Intention accuracy (%) | | | |
| --- | --- | --- | --- | --- | --- |
|  |  | LK | LC | RC | Macro avg. |
| **No attack** | Precision | 71 | 96 | 100 | 89 |
|  | Recall | 100 | 73 | 89 | 87 |
|  | F1 | 83 | 83 | 94 | 87 |
| **One-feature DE attack** | Precision | 62 | 85 | 100 | 82 |
|  | Recall | 90 | 67 | 84 | 80 |
|  | F1 | 73 | 75 | 91 | 80 (-8%) |

## 4.4 Effect of Fine-tuning Depth

For the Llama2-13B model, we fine-tune a variant by stopping the training process early (e.g., after 4500 iterations), summarize the results in Table 5 and Table 6, and compare the performance with the baseline above that stops after 7000 iterations.

Under normal conditions, the 4500-iteration model yields a slightly lower F1 score (90) compared to 92 from the baseline model, together with slightly larger RMSE in position predictions (e.g., 0.71 vs. 0.70 at the fourth second), indicating insufficient learning with reduced finetuning. However, under DE attacks, dislike the baseline model's significant reduction (a 12% drop to 81) in F1 score, the 4500-iteration model drops mildly to 84 (a 7% drop). This confirms that an LLM model finetuned to perform accurately under a normal scenario may be over-specialized with strong feature correlations, making it overly reliant on precise feature values and more vulnerable to perturbations in these features.

The findings suggest that adequate regularization techniques (e.g., stopping the fine-tuning early) could be helpful to benefit the trade-off between accuracy and robustness. In the following section, we show that incorporating CoT reasoning in the prediction task could potentially improve the accuracy without compromising robustness.

Table 5. Errors in the predicted positions with Llama2-13B model trained by 4500 iterations.

| | | RMSE of Predicted Positions (m) | | | |
| --- | --- | --- | --- | --- | --- |
| | | 1s | 2s | 3s | 4s |
| **No attack** | Lateral | 0.04 | 0.22 | 0.47 | 0.66 |
| | Longitudinal | 0.06 | 0.09 | 0.29 | 0.76 |
| | Avg. | 0.05 | 0.16 | 0.38 | 0.71 |
| **One-feature DE attack** | Lateral | 0.03 | 0.21 | 0.47 | 0.72 |
| | Longitudinal | 0.07 | 0.10 | 0.33 | 0.78 |
| | Avg. | 0.05 | 0.16 | 0.40 | 0.75 |

Table 6. Accuracy of predicted lane-change intentions using Llama2-13B model trained by 4500 iterations.

| | | Intention accuracy (%) | | | |
| --- | --- | --- | --- | --- | --- |
| | | LK | LC | RC | Macro avg. |
| **No attack** | Precision | 77 | 95 | 100 | 91 |
| | Recall | 100 | 83 | 90 | 91 |
| | F1 | 87 | 88 | 95 | 90 |
| **One-feature DE attack** | Precision | 67 | 89 | 100 | 85 |
| | Recall | 94 | 70 | 90 | 85 |
| | F1 | 78 | 78 | 95 | 84 (-7%) |

### 4.5 Role of Chain-of-Thought Reasoning

Table 7 and Table 8 examine the accuracy and robustness of the prediction model incorporating CoT reasoning. The CoT-enhanced model produces an RMSE of 0.69 when predicting the fourth-second position under normal conditions, comparable to the RMSE of 0.70 without CoT. Meanwhile, the model with CoT achieved an F1 score of 95, a 3% increase compared to 92 without CoT. The improvement could be due to the CoT model's ability to articulate reasoning steps that refine predictions.

Under attacks, the CoT-enhanced model's F1 score dropped by 7% (to 88), less than the 12% drop (to 81) for the non-CoT model. Similarly, the RMSE of position prediction also observes a limited deterioration (a 12% increase), compared with the 29% increase for the non-CoT model. This increased robustness may arise from CoT's structured reasoning, which distributes the impact of one-feature perturbation across multiple features involved in the reasoning process. Such redistribution likely dilutes the adversarial effect, compared to the strong reliance on specific, accurate feature values in the non-CoT model. The results demonstrated that incorporating CoT reasoning enhanced the robustness without sacrificing accuracy, highlighting incorporating CoT reasoning as a promising defense solution against adversarial attacks. The downside is that the fine-tuning and inference with CoT bring additional computational burden, which will be discussed in detail in Section 5.

Table 7. Errors in the predicted positions using Llama2-13B model with CoT.

|  |  | RMSE of Predicted Positions (m) | | | |
| --- | --- | --- | --- | --- | --- |
|  |  | 1s | 2s | 3s | 4s |
| No attack | Lateral | 0.03 | 0.18 | 0.36 | 0.52 |
|  | Longitudinal | 0.06 | 0.11 | 0.36 | 0.86 |
|  | Avg. | 0.05 | 0.15 | 0.36 | 0.69 |
| One-feature DE attack | Lateral | 0.03 | 0.22 | 0.47 | 0.68 |
|  | Longitudinal | 0.06 | 0.11 | 0.37 | 0.85 |
|  | Avg. | 0.05 | 0.17 | 0.42 | 0.77 (12%) |

Table 8. Accuracy of the predicted lane-change intention using Llama2-13B model with CoT.

|  |  | Intention accuracy (%) | | | |
| --- | --- | --- | --- | --- | --- |
|  |  | LK | LC | RC | Macro avg. |
| No attack | Precision | 85 | 100 | 100 | 95 |
|  | Recall | 100 | 91 | 95 | 95 |
|  | F1 | 92 | 95 | 97 | 95 |
| One-feature DE attack | Precision | 71 | 100 | 100 | 90 |
|  | Recall | 100 | 70 | 100 | 90 |
|  | F1 | 83 | 82 | 100 | 88 (-7%) |

## 4.6 Feature-level Vulnerability Analysis

The one-feature DE attack facilitates feature-level vulnerability analysis, revealing critical inputs affecting prediction performance. By systematically perturbing different features and measuring prediction performance drops, the DE attack model identifies the top vulnerable features. More specifically, for the test on each data sample, the one-feature DE attack model outputs which feature is targeted, together with the perturbation added to the feature, and the prediction performance under the perturbation. We are then able to identify the top frequently exploited features and the resultant effects by analyzing the outputs from the collection of tests.

Figure 2A shows the top vulnerable features for the 13B model without CoT. In general, it can be observed that several features are disproportionately selected and optimized by the DE algorithm, including the velocities of ahead, left front and right front vehicles, and the ego vehicle's distance to the following vehicle and left front vehicle. DE-generated perturbations to these features result in significant impacts on the prediction accuracy of the lane-change intention. This insight suggests that robustness could be improved by *prioritizing feature sanitization* or *redundancy checks for the inputs of these key features*, demonstrating the attack's utility in model hardening.

Compared with the top vulnerable features for the CoT-enhanced model (Figure 2B), we observe the impacts are less stemmed from a few features, though some of the features overlap with those in Figure 2A. This again suggests that the CoT-enhanced model distills the over-reliance on several features for prediction, consequently contributing to the model's robustness. It is worth noting that these observations are based on the attacks targeting one feature and may not be valid under cooperated attacks targeting multiple features, which will be investigated in future work.

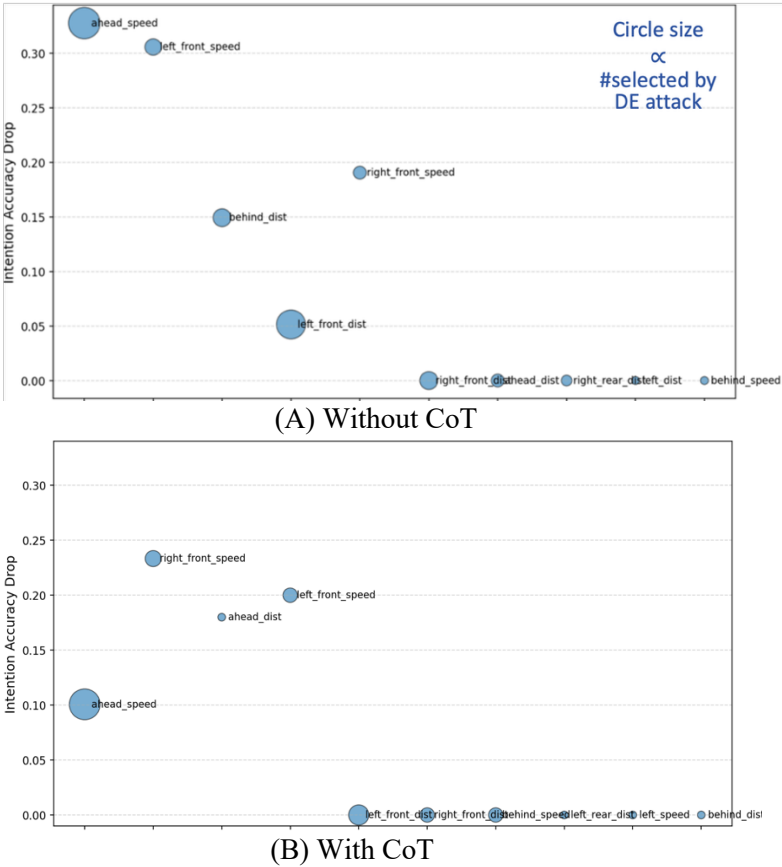

(A) Without CoT

(B) With CoT

**Figure 2**. Top vulnerable features for the 13B model. Each circle represents one feature that was selected for perturbations, with its size proportional to the counts being selected and its vertical position indicating the mean impacts of perturbing this feature.

## 5. Conclusion

This study presented the first systematic evaluation of adversarial vulnerability in LLM-based vehicle trajectory and lane-change intention prediction. By introducing a one-feature differential-evolution attack, we demonstrated that subtle, physically plausible perturbations to surrounding-vehicle features can mislead fine-tuned LLMs, resulting in degraded prediction accuracy and increased trajectory errors. Comparative experiments across model scales, training iterations, and reasoning modes suggest that larger and highly specialized models could improve the prediction performance but are likely to be more susceptible to perturbations. This indicates the critical tradeoff between accuracy and robustness in an adversarial operation environment. Our preliminary results show that incorporating Chain-of-Thought reasoning not only improves interpretability but also enhances robustness, offering a promising direction to mitigate the revealed vulnerability. The one-feature attack framework further enabled feature-level vulnerability identification, offering valuable insights for improving model safety in autonomous-driving applications.

Despite these contributions, several limitations remain, which open avenues for future research. First, the proposed DE-based attack is computationally expensive due to the iterative optimization and repeated model querying required for black-box evaluation, which is a challenge shared broadly within the adversarial learning community. Future work could explore efficiency improvements by employing specialized lightweight LLM architectures, developing transferable perturbations across models and datasets, or constructing pre-computed lookup tables to accelerate vulnerability screening (Demontis et al., 2019; X. Wang et al., 2024). Second, while this study focused on LLM-based predictors, it did not include comparisons with other state-of-the-art spatiotemporal models, such as graph convolutional network–long short-term memory (GCN-LSTM) hybrids (J. Wang et al., 2024). Future experiments should systematically benchmark LLMs against these alternatives to determine whether the advantages of LLMs in accuracy offset their susceptibility to adversarial manipulation. Third, our attack framework targeted a single feature of one surrounding vehicle to maintain realism and stealthiness, yet coordinated multi-feature or multi-agent perturbations could produce even more severe disruptions, which could be an important direction for understanding compound vulnerabilities in connected traffic systems. Finally, this work primarily focused on attack modeling and vulnerability analysis, while defenses remain underexplored. Future research should therefore emphasize developing robust mitigation strategies, such as adversarial fine-tuning, input verification, or CoT-guided defense mechanisms, to enhance the resilience of LLM-based transportation systems in safety-critical scenarios.

## Acknowledgement

The authors would like to express their sincere gratitude to Mingxing Peng of the Hong Kong University of Science and Technology (Guangzhou) for generously sharing the data and code from her previous work, as well as for her valuable discussions. The authors also thank Jinzhou Cao at Shenzhen Technology University, Zhanbo Sun and Yueyuan Du at Southwest Jiaotong University for their valuable insights in interpreting the results and improving figures in the manuscript. The authors acknowledge the use of ChatGPT (OpenAI) for grammar checking and language refinement during manuscript preparation. All AI-assisted revisions were thoroughly reviewed and validated by the authors.

# Appendix

## A. Details of LLM Baseline

As noted in Section 4.1, we finetune Llama-2-13B model open-sourced by Meta (Touvron et al., 2023) as the baseline. The key parameter settings are listed in Table 9. The fine-tuning uses DeepSpeed on a server with two NVIDIA RTX3090 GPUs.

Table 9. Key parameters for fine-tuning the LLM.

| Parameter | Values |
| --- | --- |
| Learning rate | 5e-4 |
| Batch size | 8 |
| Training epochs | 2 |
| LoRA rank | 64 |
| LoRA alpha | 16 |
| Gradient accumulation steps | 8 |
| Warmup steps | 600 |

Some examples of the prompts for fine-tuning and running the LLM are provided below. Interested readers are referred to (Peng et al., 2025) for details of preparing the prompts using the HighD dataset.

### A.1 Prompt Templates

The input prompts consist of a fixed system message and a scenario-specific user message.

**System Message**

Role: You are an expert driving prediction model of an autonomous driving system, that can predict the future driving intention and future 4-second driving trajectory for a given ego vehicle, avoiding collision with other vehicles and obstacles on the road. Context:
- Coordinates:
  Y-axis is perpendicular, and X-axis is parallel to the direction ego vehicle is facing. Ego vehicle's current position is (0,0). Positive values on the y-axis represent the left side of the ego vehicle, and negative values on the y-axis represent the right side of the vehicle.
- Output:
  - Final Answer:
    Intention: 0 (Keep lane), 1 (Left lane change), 2 (Right lane change). The final answer should be one of the three modes.

**User Message Example**

The ego vehicle is driving on a four-lane highway, located at the rightmost lane.
- The information of ego vehicle is as follow:
  - Velocity(km/h): vx=78.52, vy=2.05;
  - Acceleration(m/s^2): ax=2.20, ay=-2.20;
  - Type: Car, with width of 2.02 m and length of 4.65 m;
  - Historical position of the last 2 seconds (One point every 0.4s): [(-41.30,-0.79), (-33.92,-0.71), (-25.71,-0.58), (-17.37,-0.41), (-8.79,-0.22), (0.0,0.0)].
- The information of its surrounding vehicles (with a range of 200m) are listed as follow:
  - Left front: a Car traveling at 85.43 km/h of X-axis, with a distance of 103 m;
  - Left rear: a Car traveling at 91.84 km/h of X-axis, with a distance of 60 m.

### A.2 Example of LLM Response

> **Final Answer:**
> Intention: 1: Left lane change;
> Trajectory: [(22.07,0.59), (44.45,1.08), (66.97,1.42), (89.56,1.61)].

Consider the above user message prompt. The fine-tuned LLM generates:

This demonstrates how the model uses CoT to reason about features like lateral velocity and surrounding space before predicting.

### A.3 Model with CoT reasoning

To account for the CoT reasoning, the prompts used for fine-tuning the LLM need to contain additional messages. More specifically, the system message specifies the requirement of giving the reasoning of notable features leading to the answer and the potential behaviors (i.e., Thought: Notable features; Potential behaviors). And the response of LLM provides additional information:

> **- Thought:**
> Notable features: vy = 2.05;
> Notable features: ax = 2.20;
> Notable feature: Left front is free;
> Potential behavior: Change left to the fast lane.

### B. Effect of Perturbation Budget

As noted in Section 3.2.3, we test various perturbation budgets (i.e., values of Δ) to investigate the sensitivity to perturbation magnitude. With a 0.1 bound, the F1 score dropped by 12% (to 81). At 0.3, the F1 score fell by 22% (to 72). This indicates that larger perturbations amplify error propagation through the LLM's attention and decoding layers, likely due to the sensitivity of transformer layers that could amplify errors in input data. The findings emphasize the need for tight input validation in LLM-based AV deployments.

**Table 10.** Errors in the predicted positions with various perturbation budgets Δ (13B model without CoT).

| Perturbation budgets | | RMSE of Predicted Positions (m) | | | |
|---|---|---|---|---|---|
| | | 1s | 2s | 3s | 4s |
| 0.1 | Lateral | 0.04 | 0.24 | 0.56 | 0.93 |
| | Longitudinal | 0.05 | 0.11 | 0.36 | 0.86 |
| | Avg. | 0.05 | 0.18 | 0.46 | 0.90 |
| 0.2 | Lateral | 0.04 | 0.25 | 0.57 | 0.89 |
| | Longitudinal | 0.06 | 0.12 | 0.41 | 0.99 |
| | Avg. | 0.05 | 0.19 | 0.49 | 0.94 |
| 0.3 | Lateral | 0.04 | 0.28 | 0.73 | 1.25 |
| | Longitudinal | 0.06 | 0.10 | 0.39 | 1.01 |
| | Avg. | 0.05 | 0.19 | 0.56 | 1.13 |

**Table 11**. Accuracy of the predicted lane-change intention with various perturbation budgets Δ (13B model without CoT).

| Perturbation budgets | | Intention accuracy (%) | | | |
|---|---|---|---|---|---|
| | | LK | LC | RC | Macro avg. |
| **0.1** | P | 64 | 81 | 96 | 80 |
| | R | 76 | 75 | 89 | 80 |
| | *F1* | *70* | *78* | *93* | *81* |
| **0.2** | P | 60 | 78 | 83 | 74 |
| | R | 55 | 78 | 91 | 75 |
| | *F1* | *57* | *78* | *87* | *74* |
| **0.3** | P | 67 | 67 | 90 | 74 |
| | R | 38 | 83 | 96 | 72 |
| | *F1* | *48* | *74* | *93* | *72* |

## Reference

Akhtar, N., Mian, A., 2018. Threat of Adversarial Attacks on Deep Learning in Computer Vision: A Survey. IEEE Access 6, 14410–14430. https://doi.org/10.1109/ACCESS.2018.2807385

Biggio, B., Roli, F., 2018. Wild patterns: Ten years after the rise of adversarial machine learning. Pattern Recognition 84, 317–331. https://doi.org/10.1016/j.patcog.2018.07.023

Cai, X., Liu, D., Qu, X., Fang, X., Dong, J., Tang, K., Zhou, P., Sun, L., Hu, W., 2025. Towards Building Model/Prompt-Transferable Attackers against Large Vision-Language Models. Presented at the 39th Conference on Neural Information Processing Systems (NeurIPS 2025).

Cao, Y., Xiao, C., Cyr, B., Zhou, Y., Park, W., Rampazzi, S., Chen, Q.A., Fu, K., Mao, Z.M., 2019. Adversarial Sensor Attack on LiDAR-based Perception in Autonomous Driving. Proceedings of the 2019 ACM SIGSAC Conference on Computer and Communications Security 2267–2281. https://doi.org/10.1145/3319535.3339815

Chao, P., Robey, A., Dobriban, E., Hassani, H., Pappas, G.J., Wong, E., 2025. Jailbreaking Black Box Large Language Models in Twenty Queries, in: 2025 IEEE Conference on Secure and Trustworthy Machine Learning (SaTML). pp. 23–42. https://doi.org/10.1109/SaTML64287.2025.00010

Choudhary, T., Dewangan, V., Chandhok, S., Priyadarshan, S., Jain, A., Singh, A.K., Srivastava, S., Jatavallabhula, K.M., Krishna, K.M., 2024. Talk2bev: Language-enhanced bird's-eye view maps for autonomous driving. Presented at the 2024 IEEE International Conference on Robotics and Automation (ICRA), IEEE, pp. 16345–16352.

Chung, N., Gao, S., Vu, T.-A., Zhang, J., Liu, A., Lin, Y., Dong, J.S., Guo, Q., 2024. Towards Transferable Attacks Against Vision-LLMs in Autonomous Driving with Typography. https://doi.org/10.48550/arXiv.2405.14169

Clusmann, J., Ferber, D., Wiest, I.C., Schneider, C.V., Brinker, T.J., Foersch, S., Truhn, D., Kather, J.N., 2025. Prompt injection attacks on vision language models in oncology. Nature Communications 16, 1239. https://doi.org/10.1038/s41467-024-55631-x

Cui, Y., Huang, S., Zhong, J., Liu, Z., Wang, Y., Sun, C., Li, B., Wang, X., Khajepour, A., 2023. Drivellm: Charting the path toward full autonomous driving with large language models. IEEE Transactions on Intelligent Vehicles 9, 1450–1464.

Demontis, A., Melis, M., Pintor, M., Jagielski, M., Biggio, B., Oprea, A., Nita-Rotaru, C., Roli, F., 2019. Why Do Adversarial Attacks Transfer? Explaining Transferability of Evasion and Poisoning Attacks. Presented at the 28th USENIX Security Symposium (USENIX Security 19), pp. 321–338.

Deng, Y., Tu, Z., Yao, J., Zhang, M., Zhang, T., Zheng, X., 2025. Target: Traffic rule-based test generation for autonomous driving systems. IEEE Transactions on Software Engineering.

Goodfellow, I.J., Shlens, J., Szegedy, C., 2014. Explaining and harnessing adversarial examples. arXiv preprint arXiv:1412.6572.
Guo, X., Zhang, Q., Jiang, J., Peng, M., Zhu, M., Yang, H.F., 2024. Towards explainable traffic flow prediction with large language models. Communications in Transportation Research 4, 100150. https://doi.org/10.1016/j.commtr.2024.100150
Hall, M., van der Maaten, L., Gustafson, L., Jones, M., Adcock, A., 2022. A systematic study of bias amplification. arXiv preprint arXiv:2201.11706.
Huang, Y., Gao, C., Wu, S., Wang, H., Wang, X., Zhou, Y., Wang, Y., Ye, J., Shi, J., Zhang, Q., Li, Yuan, Bao, H., Liu, Z., Guan, T., Chen, D., Chen, R., Guo, K., Zou, A., Kuen-Yew, B.H., Xiong, C., Stengel-Eskin, E., Zhang, Hongyang, Yin, H., Zhang, Huan, Yao, H., Yoon, J., Zhang, J., Shu, K., Zhu, K., Krishna, R., Swayamdipta, S., Shi, T., Shi, W., Li, X., Li, Yiwei, Hao, Y., Hao, Y., Jia, Z., Li, Z., Chen, X., Tu, Z., Hu, X., Zhou, T., Zhao, J., Sun, L., Huang, F., Sasson, O.C., Sattigeri, P., Reuel, A., Lamparth, M., Zhao, Y., Dziri, N., Su, Y., Sun, H., Ji, H., Xiao, C., Bansal, M., Chawla, N.V., Pei, J., Gao, J., Backes, M., Yu, P.S., Gong, N.Z., Chen, P.-Y., Li, B., Zhang, X., 2025. On the Trustworthiness of Generative Foundation Models: Guideline, Assessment, and Perspective. https://doi.org/10.48550/arXiv.2502.14296
Jiang, W., Li, H., Liu, S., Luo, X., Lu, R., 2020. Poisoning and Evasion Attacks Against Deep Learning Algorithms in Autonomous Vehicles. IEEE Trans. Veh. Technol. 69, 4439–4449. https://doi.org/10.1109/TVT.2020.2977378
Kaur, D., Uslu, S., Rittichier, K.J., Durresi, A., 2023. Trustworthy Artificial Intelligence: A Review. ACM Comput. Surv. 55, 1–38. https://doi.org/10.1145/3491209
Krajewski, R., Bock, J., Kloeker, L., Eckstein, L., 2018. The highD Dataset: A Drone Dataset of Naturalistic Vehicle Trajectories on German Highways for Validation of Highly Automated Driving Systems, in: 2018 21st International Conference on Intelligent Transportation Systems (ITSC). Presented at the 2018 21st International Conference on Intelligent Transportation Systems (ITSC), pp. 2118–2125. https://doi.org/10.1109/ITSC.2018.8569552
Liu, F., Liu, H., Jiang, W., 2022. Practical adversarial attacks on spatiotemporal traffic forecasting models. Advances in Neural Information Processing Systems 35, 19035–19047.
Liu, Yi, Deng, G., Li, Y., Wang, K., Wang, Z., Wang, X., Zhang, T., Liu, Yepang, Wang, H., Zheng, Y., Liu, Yang, 2024. Prompt Injection attack against LLM-integrated Applications. https://doi.org/10.48550/arXiv.2306.05499
Ludan, J.M., Meng, Y., Nguyen, T., Shah, S., Lyu, Q., Apidianaki, M., Callison-Burch, C., 2023. Explanation-based finetuning makes models more robust to spurious cues.
Madry, A., Makelov, A., Schmidt, L., Tsipras, D., Vladu, A., 2017. Towards deep learning models resistant to adversarial attacks. arXiv preprint arXiv:1706.06083.
Mao, J., Qian, Y., Ye, J., Zhao, H., Wang, Y., 2023. GPT-Driver: Learning to Drive with GPT. https://doi.org/10.48550/arXiv.2310.01415
Mei, Y., Nie, T., Sun, J., Tian, Y., 2025. Llm-attacker: Enhancing closed-loop adversarial scenario generation for autonomous driving with large language models. arXiv preprint arXiv:2501.15850.
Movahedi, M., Choi, J., 2024. The crossroads of llm and traffic control: A study on large language models in adaptive traffic signal control. IEEE Transactions on Intelligent Transportation Systems.
Peng, M., Guo, X., Chen, X., Chen, K., Zhu, M., Chen, L., Wang, F.-Y., 2025. LC-LLM: Explainable lane-change intention and trajectory predictions with Large Language Models. Communications in Transportation Research 5, 100170. https://doi.org/10.1016/j.commtr.2025.100170
Petit, J., Shladover, S.E., 2015. Potential Cyberattacks on Automated Vehicles. IEEE Transactions on Intelligent Transportation Systems 16, 546–556. https://doi.org/10.1109/TITS.2014.2342271
Souly, A., Rando, J., Chapman, E., Davies, X., Hasircioglu, B., Shereen, E., Mougan, C., Mavroudis, V., Jones, E., Hicks, C., 2025. Poisoning Attacks on LLMs Require a Near-constant Number of Poison Samples. arXiv preprint arXiv:2510.07192.
Storn, R., Price, K., 1997. Differential evolution–a simple and efficient heuristic for global optimization over continuous spaces. Journal of global optimization 11, 341–359.
Su, J., Vargas, D.V., Kouichi, S., 2019. One pixel attack for fooling deep neural networks. https://doi.org/10.1109/TEVC.2019.2890858

Touvron, H., Martin, L., Stone, K., Albert, P., Almahairi, A., Babaei, Y., Bashlykov, N., Batra, S., Bhargava, P., Bhosale, S., 2023. Llama 2: Open foundation and fine-tuned chat models. arXiv preprint arXiv:2307.09288.

Wandelt, S., Zheng, C., Wang, S., Liu, Y., Sun, X., 2024. Large Language Models for Intelligent Transportation: A Review of the State of the Art and Challenges. Applied Sciences 14. https://doi.org/10.3390/app14177455

Wang, F., Wang, X., Ban, X. (Jeff), 2024. Data poisoning attacks in intelligent transportation systems: A survey. Transportation Research Part C: Emerging Technologies 165, 104750. https://doi.org/10.1016/j.trc.2024.104750

Wang, J., Liu, K., Li, H., 2024. LSTM-based graph attention network for vehicle trajectory prediction. Computer Networks 248, 110477. https://doi.org/10.1016/j.comnet.2024.110477

Wang, X., Wang, F., Hong, Y., Ban, X., 2024. Transferability in Data Poisoning Attacks on Spatiotemporal Traffic Forecasting Models.

Wen, L., Fu, D., Li, X., Cai, X., Ma, T., Cai, P., Dou, M., Shi, B., He, L., Qiao, Y., 2023. Dilu: A knowledge-driven approach to autonomous driving with large language models. arXiv preprint arXiv:2309.16292.

Wu, K., Li, P., Zhou, Y., Gan, R., You, J., Cheng, Y., Zhu, J., Parker, S.T., Ran, B., Noyce, D.A., 2025. V2x-llm: Enhancing v2x integration and understanding in connected vehicle corridors. arXiv preprint arXiv:2503.02239.

Yang, F., Liu, X.C., Lu, L., Wang, B., Liu, C., 2025. A Self-Supervised Multi-Agent Large Language Model Framework for Customized Traffic Mobility Analysis Using Machine Learning Models. Transportation Research Record 2679, 1–16. https://doi.org/10.1177/03611981251322468

Yang, W., Bi, X., Lin, Y., Chen, S., Zhou, J., Sun, X., 2024. Watch out for your agents! investigating backdoor threats to llm-based agents. Advances in Neural Information Processing Systems 37, 100938–100964.

Zhang, W., Kong, X., Dewitt, C., Braunl, T., Hong, J.B., 2024. A study on prompt injection attack against llm-integrated mobile robotic systems. Presented at the 2024 IEEE 35th International Symposium on Software Reliability Engineering Workshops (ISSREW), IEEE, pp. 361–368.

Zhao, H., Chen, H., Yang, F., Liu, N., Deng, H., Cai, H., Wang, S., Yin, D., Du, M., 2024. Explainability for Large Language Models: A Survey. ACM Trans. Intell. Syst. Technol. 15. https://doi.org/10.1145/3639372

Zhao, K., Li, L., Ding, K., Gong, N.Z., Zhao, Y., Dong, Y., 2025. A survey on model extraction attacks and defenses for large language models. Presented at the Proceedings of the 31st ACM SIGKDD Conference on Knowledge Discovery and Data Mining V. 2, pp. 6227–6236.

Zhao, Y., Wang, P., Zhao, Yibo, Du, H., Yang, H.F., 2025. SafeTraffic Copilot: adapting large language models for trustworthy traffic safety assessments and decision interventions. Nature Communications 16, 8846. https://doi.org/10.1038/s41467-025-64574-w

Zou, A., Wang, Z., Carlini, N., Nasr, M., Kolter, J.Z., Fredrikson, M., 2023. Universal and transferable adversarial attacks on aligned language models. arXiv preprint arXiv:2307.15043.